# Context- and Sequence-Aware Convolutional Recurrent Encoder for Neural Machine Translation

Ritam Mallick
Delhi Technological University
Rohini, Delhi-42
India
ritammallick.mallick9@gmail.com

Seba Susan
Delhi Technological University
Rohini, Delhi-42
India
seba_406@yahoo.in

Vaibhaw Agrawal
Delhi Technological University
Rohini, Delhi-42
India
vaibhaw2731@yahoo.com

Rizul Garg
Delhi Technological University
Rohini, Delhi-42
India
gargrizul@gmail.com

Prateek Rawal
Delhi Technological University
Rohini, Delhi-42
India
prateekrawal11@gmail.com

## ABSTRACT

Neural Machine Translation model is a sequence-to-sequence converter based on neural networks. Existing models use recurrent neural networks to construct both the encoder and decoder modules. In alternative research, the recurrent networks were substituted by convolutional neural networks for capturing the syntactic structure in the input sentence and decreasing the processing time. We incorporate the goodness of both approaches by proposing a convolutional-recurrent encoder for capturing the context information as well as the sequential information from the source sentence. Word embedding and position embedding of the source sentence is performed prior to the convolutional encoding layer which is basically a n-gram feature extractor capturing phrase-level context information. The rectified output of the convolutional encoding layer is added to the original embedding vector, and the sum is normalized by layer normalization. The normalized output is given as a sequential input to the recurrent encoding layer that captures the temporal information in the sequence. For the decoder, we use the attention-based recurrent neural network. Translation task on the German-English dataset verifies the efficacy of the proposed approach from the higher BLEU scores achieved as compared to the state of the art.

## CCS CONCEPTS

• **Computing methodologies**~Artificial intelligence~Natural language processing~Machine translation

## KEYWORDS

Neural machine translation, convolutional, recurrent, encoder, decoder, context, sequence-to-sequence model

## 1 INTRODUCTION

Neural Machine Translation (NMT) model is a sequence-to sequence converter that translates a variable length source sentence to a variable length target sentence using only neural networks. The precursor of the current state of the art in NMT can be attributed to Kalchbrenner and Blunsom (2013) [1] and Sutskever *et al.* (2014) [2]; both works condition the probability of each word in the target sentence based on the source sentence representation. Separate recurrent neural networks were used in [2] for processing the source sentence and for predicting the target sentence. The problem with these encoder-decoder models, as they are called [3], is the drop in performance with increase in the length of the source sentence and the presence of unknown words [4]. Bahdanau *et al.* proposed a solution involving a soft-search through sentences to find a suitable segment of the source sentence that can be translated effectively [5]. His approach, called the attention mechanism, is popularly incorporated into current NMT models [8, 17]. An alternative to recurrent neural network is the convolutional neural network (CNN) [1]. Unlike recurrent networks, CNN enables parallelization and faster processing. Encoder-decoder models using CNN were proved effective in translating phrases in the source sentence to suitable target sentences [6, 7]. CNN based NMT models could not, however, match the performance of the state of the art in recurrent neural network based NMT models [3]. Our work integrates convolutional and recurrent layers for constructing the NMT encoder, in order to extract both the context and the temporal information in the source sentence. The organization of this paper is as follows. Section 2 presents the proposed approach. Section 3 discusses the experimental setup section 4 analyzes the results of the translation task. Section 5 summarizes the paper.

## 2 PROPOSED MODEL

A few approaches, in the past, have combined convolutional and recurrent architectures for NMT. Quasi-RNN (QRNN) proposed by Bradbury *et al.* [9] alternates convolutional layers with the recurrent pooling function that is non-trainable. Its translation performance is lower than the attention-based models which our model outperforms. Another example is the Convolution over Recurrent model (CoveR) [10] in which convolutional layers are added at the output of the RNN encoder to capture the context information. In our work, we propose a novel convolutional-cum-recurrent encoder (shown in Fig.1) to combine the goodness of both approaches. The resultant encoder is both context-aware and sequence-aware. Bahdanau *et al.*'s RNN with attention mechanism [5] constitutes the decoder module in our NMT system. All layers are fully trainable.

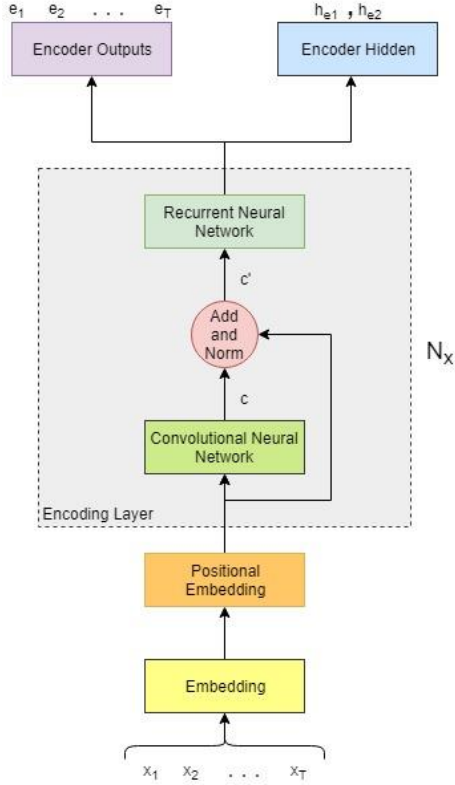

**Figure 1: Proposed Convolutional Recurrent encoder model**

### 3.1 Problem definition

Consider that the source sentence of length $T$ is denoted by the sequence $\{x_1,...,x_T\}$ and the translated target sequence of length $M$ is denoted by $\{y_1,...,y_M\}$. For the sentence pair (**x**, **y**) the task is to learn a probability distribution $p(y_i|x_1,...,x_T,y_{1:i-1})$ that predicts the next target word $y_i$, given the partial translation $\{y_1,...,y_{i-1}\}$ and the encoded source sentence. The probability of the target sentence **y** given a source sentence **x**, is given by

$$p(\mathbf{y}|\mathbf{x}) = \prod_{i=1}^{M} p(y_i|x_1,...,x_T,y_{1:i-1}) \qquad (1)$$

### 3.2 Encoder

A novel NMT encoder architecture is proposed in our paper that comprises of the convolutional neural network in the first subsection followed by the recurrent neural network in the second subsection. All layers are fully trainable. The CNN and the RNN subsections of the encoder are explained next in more detail.

#### 3.2.1 CNN subsection

Akin to the work of Gehring *et al.* on convolutional encoders [6], in a bid to generate position-aware encoding vectors, the word embedding **l** is augmented (summed up) with position embedding **p** [14] to record the position of each token in the source sentence. Let $a_i \in \mathbb{R}^{1 \times d}$ be the $d$-dimensional augmented embedding for the $i^{\text{th}}$ discrete token ($d$=512), that is given as input to the convolutional network. The advantage of using the convolutional layers first is the n-gram feature extraction that captures context by applying a filter of size $n$ on the embedding ensuing from the source sentence **x**. Multiple layers are vertically stacked to capture context from longer sentences, as shown in Fig. 2.

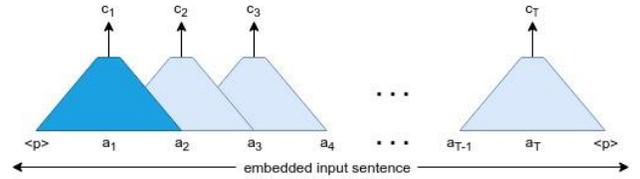

**Figure 2: Stacking multiple convolutional layers to capture context**

To facilitate efficient learning, we use skip connections [11] from the input of the convolutional layer to its output. Each convolutional layer is followed by a non-linearity (*tan h* function). The rectified convolutional encoded vector $c$ is summed up with the word and position embedding through a residual connection, and normalized by layer normalization procedure [12]. The normalized sequential stream of vectors $c'$, corresponding to the input tokens in the source sentence, are fed as input to the recurrent part of the encoder where these are encoded as hidden state vectors that constitute the input to the decoder module.

#### 3.2.2 RNN subsection

Recurrent neural networks are used to encase the temporal sequence of words in the source sentence that are represented by word embeddings, a significant advance over the bag-of-words representation [15]. The temporal pattern of the encoded sequence is learnt by the RNN subsection shown in Fig. 3. The normalized output of the CNN part of the encoder is fed as input to the bidirectional recurrent network which forms the second subsection of our encoder module. There are two RNN encoders, $e_1$ for processing the input sentence in the forward direction and the encoder $e_2$ that processes the sentence in the reverse direction. The output of the bidirectional RNN is the concatenation of the forward pass and backward pass outputs at each time-step $t$.

$$e_t = \left[ \vec{e_t} : \overleftarrow{e_t} \right] \qquad (2)$$

The encoder output is the annotation obtained for the input source sentence, which is passed as the input to the RNN decoder. The forward pass output is computed as a function of the previous hidden state and the convolutional input.

$$\overrightarrow{e_t} = f\left(c_t', h_{e_1}^{<t-1>}\right) \quad (3)$$

Likewise, the backward pass output is computed.

$$\overleftarrow{e_t} = f\left(c_t', h_{e_2}^{<t+1>}\right) \quad (4)$$

Here, $f$ is a non-linear activation function. We use Gated Recurrent Unit (GRU) [3] for implementing the non-linear recurrent function, which is a simplified version of the Long Short Term Memory (LSTM) [13].

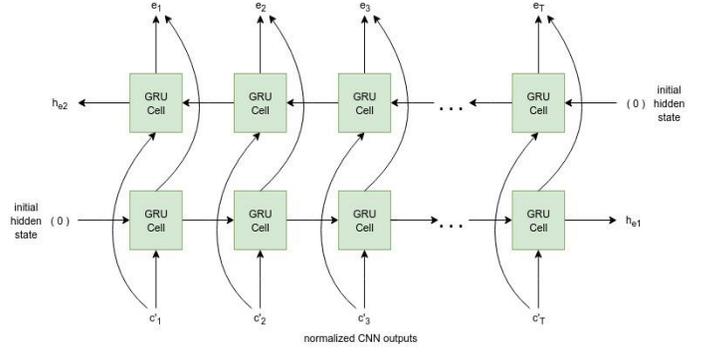

Figure 3: RNN subsection of the encoder

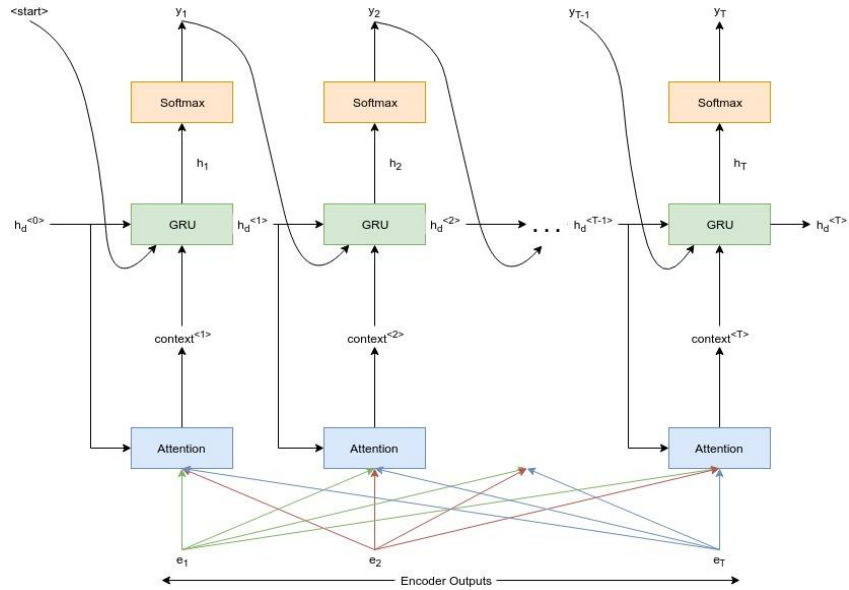

Figure 4: Decoder architecture

### 3.3 Decoder

Our decoder module (shown in Fig. 4) comprises of a recurrent neural network with attention mechanism [5] that computes a probability distribution for every word in the target sentence. The probability of the output target word is computed as a function of the previously predicted word and the context vector. A soft weight mechanism is used to evaluate the context vector from the encoder output. The context vector at time-step $i$, is defined as the weighted sum of the encoder hidden states computed in (2).

$$context^{<i>} = \sum_{t=1}^{T} \alpha_{it} e_t \quad (5)$$

The soft attention weights are computed over the entire source sentence of length $T$ as

$$\alpha_{it} = \frac{e^{score(h_d^{<i-1>}, e_t)}}{\sum_{h=1}^{T} e^{score(h_d^{<i-1>}, e_h)}} \quad (6)$$

$$score\left(h_d^{<i-1>}, e_t\right) = \tanh\left(W_1 h_d^{<i-1>} + b_1 + W_2 e_t + b_2\right) \quad (7)$$

To compute the probability distribution of the target sentence given the source sentence as shown in (1), a word-by-word prediction pattern is followed. The probability of the next target word is computed as a function of the previously predicted word and the context vector as

$$p\left(y_i | x_1, ..., x_T, y_{1:i-1}\right) = soft\max(g(h_j)) \quad (8)$$

where,

$$h_j = f(h_{j-1}, [context^{<j>} : y_{j-1}]) \quad (9)$$

## 4 RESULTS

Our experiments were conducted on 176,692 sentence pairs of the German-English dataset (*Tatoeba project*) [16]. 163,957 sentence pairs were used for training and 3900 sentence pairs for testing. 5% of the data was used for validation, that served the purpose of tuning of hyper-parameters such as the size of the embedding vector, number of hidden units and number of encoding layers. The batch size was set to 128 sentence pairs. We compare our model, implemented using Python on a Tesla K80 GPU, with that of Kalchbrenner and Blunsom (2013) [1] that used a CNN for the encoder and a RNN for the decoder, that we call as CNN-1. The next work we compare with, is the pure RNN model of Bahdanau *et al.* (2014) [5] that incorporates attention mechanism. We also compare our work with the convolutional NMT model of Gehring *et al.* (2016) [7], referred to as CNN-2. The results are compiled in Table 1. The presence or absence of position embedding and attention mechanism in these models is also indicated. We conducted an empirical study on the variation of BLEU scores when the number of stacked convolutional encoding layers is varied from 1 to 5. The highest BLEU score of 30.6 is obtained when $x$=3. For the encoder and decoder RNNs, the Adadelta optimizer is used with learning rate 0.1 and threshold 1e-6. We find that our convolutional-cum-recurrent encoder model yields the highest BLEU scores of 30.6 with position embedding and 27.9 without position embedding. In comparison, the purely CNN model of CNN-2 gives the second-best performance.

## 5 CONCLUSION

In our work, we seek to improve the RNN encoder-decoder NMT model by incorporating encoding by convolutional neural networks in the first part of the encoder to capture the syntactic structure in the source sentence. The CNN-encoded output is embossed with the embedding vector and the result is given as input to the RNN part of the encoder. The RNN hidden states are used to compute the context vector using soft attention, which helps the decoder RNN in translating the target sentence by computing the probability for each target word given the previous one. Our NMT model outperforms existing models for experiments based on the German-English dataset. Adaptation of our model to Indic languages forms the next phase of our project.

Table 1: Comparison of our model with best performing models

| Models | Encoder | | | Decoder | | BLEU |
|---|---|---|---|---|---|---|
| | Position embedding | CNN | RNN | Attention | RNN | |
| CNN-1 | No | Yes | No | No | Yes | 22.5 |
| RNN | No | No | Yes | Yes | Yes | 26.3 |
| CNN-2 | Yes | Yes | No | Yes | Yes | 29.2 |
| **Ours** | Yes | Yes | Yes | Yes | Yes | 30.6 |